\documentclass[conference]{IEEEtran}

\usepackage{bobbystyle_IEEEjournal}

\usepackage{tikz}
\usetikzlibrary{spy}

\newcommand{\dquotes}[1]{``#1''}

\usepackage{chngcntr}
\counterwithout{equation}{section}

\newcommand{\specialcell}[2][c]{%
  \begin{tabular}[#1]{@{}c@{}}#2\end{tabular}}
  
  \tikzstyle{only in spy node magn 1.75}=[%
   transform canvas={%
      shift={(tikzspyinnode)},
      scale=1.75,
   }
]

\IEEEoverridecommandlockouts

\begin{document}

\title{\fontsize{23.4}{23.4}\selectfont Deep BCD-Net Using Identical Encoding-Decoding CNN Structures for Iterative Image Recovery}

\author{\IEEEauthorblockN{Il Yong Chun\IEEEauthorrefmark{1} and Jeffrey A. Fessler\IEEEauthorrefmark{2}}
\IEEEauthorblockA{Department of Electrical Engineering and Computer Science, The University of Michigan \\
Ann Arbor, MI 48019-2122 USA\\
Email: \IEEEauthorrefmark{1}iychun@umich.edu, \IEEEauthorrefmark{2}fessler@umich.edu}
\vspace{-1.2\baselineskip}
\thanks{This work is supported in part by the Keck Foundation and NIH Grant U01 EB018753.}
}

\maketitle

\begin{abstract}
In \dquotes{extreme} computational imaging that collects extremely undersampled or noisy measurements, obtaining an accurate image within a reasonable computing time is challenging.
Incorporating image mapping convolutional neural networks (CNN) into iterative image recovery has great potential to resolve this issue.
This paper \textit{1)} incorporates image mapping CNN using identical convolutional kernels in both encoders and decoders into a block coordinate descent (BCD) signal recovery method and \textit{2)} applies alternating direction method of multipliers to train the aforementioned image mapping CNN. 
We refer to the proposed recurrent network as \emph{BCD-Net using identical encoding-decoding CNN structures}.
Numerical experiments show that, for \textit{a)} denoising low signal-to-noise-ratio images and \textit{b)} extremely undersampled magnetic resonance imaging, the proposed BCD-Net achieves significantly more accurate image recovery, compared to BCD-Net using distinct encoding-decoding structures and/or the conventional image recovery model using both wavelets and total variation.
\end{abstract}

\section{Introduction}

Using learned convolutional operators for iterative signal/image recovery is a growing trend in computational imaging \cite{Zeiler&etal:10CVPR, Heide&eta:15CVPR, Serrano&etal:16CGF, Chun&Fessler:18TIP, Chun&Fessler:17SAMPTA, Chun&Fessler:18arXiv}, improving signal recovery performances over conventional non-trained regularizers (e.g., sparsity promoting regularizers) \cite{Chun&Fessler:18TIP, Chun&Fessler:18arXiv, Chun&Fessler:17SAMPTA}. 
The iterative image recovery approaches that use learned convolutional operators or convolutional neural network (CNN) closely relate to challenging (nonconvex) block optimization. 
The authors in \cite{Chun&Fessler:18TIP, Chun&Fessler:18arXiv, Chun&Fessler:17SAMPTA} proposed a fast and convergence-guaranteed \textit{block proximal gradient method using a majorizer} to quickly and stably recover images with such image recovery approaches.
Nonetheless, the corresponding iterative algorithm needs several hundreds of iterations to converge, detracting from its practical use.

By unfolding iterative signal recovery algorithms, there exist several works in combining neural network approaches into them \cite{Gregor&Lecun:10ICML, Yang&etal:16NIPS, Kamilov&Mansour:16SPL, Zuo&etal:16TIP, Chen&Pock:17PAMI, Borgerding&Schniter&Rangan:17TSP, Hammernik&etal:17MRM, Ravishankar&Chun&Fessler:17Asilomar}. 
By optimizing image mapping networks---consisting of encoding and decoding kernels, thresholding operators, etc.---at each iteration (or layer), the methods moderate the aforementioned convergence issue, aiming to give \dquotes{best} signal estimates at each layer.
The authors in \cite{Ravishankar&Chun&Fessler:17Asilomar} incorporated iteration-wisely optimized image mapping networks into block coordinate descent (BCD) optimization method; referred to BCD-Net. 
However, encoding filters do not sufficiently capture rich information of training data (i.e., during training these filters remain close to their initial conditions) \cite{Ravishankar&Chun&Fessler:17Asilomar}, and this can limit the signal recovery performance of BCD-Net.

This paper \textit{1)} proposes a new BCD-Net using image mapping CNNs that use identical convolutional kernels in both encoders and decoders---we refer to this as the identical encoding-decoding CNN structure---and \textit{2)} applies alternating direction method of multipliers (ADMM \cite{Boyd&Parikh&Chu&Peleato&Eckstein:11FTML}) to train the proposed BCD-Net. 
Numerical experiments show that, for \textit{a)} denoising low signal-to-noise-ratio (SNR) images and \textit{b)} extremely undersampled magnetic resonance imaging (MRI), the proposed BCD-Net significantly improves image recovery accuracy compared to BCD-Net using the distinct encoding-decoding structure \cite{Ravishankar&Chun&Fessler:17Asilomar} and/or the image recovery model using both wavelets and total variation (TV) (e.g., \cite{Lustig&Dohono&Pauly:09MRM}).

\begin{algorithm}[pt!]
\caption{BCD-Net}
\label{alg:BCD-Net}

\begin{algorithmic}
\REQUIRE $\{ \mathrm{Mapping}^{(i)} : i=1,\ldots,N_{\text{Nets}}\}$, $\mb{x}^{(0)}$, $\mb{y}$, $\lambda >  0$

\FOR{$i = 0,\ldots,N_{\text{Nets}}\!-\!1$}

\STATE $\displaystyle \mb{z}^{(i+1)} = \mathrm{Mapping}^{(i+1)} \big( \mb{x}^{(i)} \big)$

\STATE $\displaystyle \mb{x}^{(i+1)} = \argmin_{\mb{x} \in \bbC^N} f(\mb{x}; \mb{y}) +  \lambda \big\| \mb{x} - \mb{z}^{(i+1)} \big\|_2^2$

\ENDFOR

\end{algorithmic}
\end{algorithm}

\section{BCD-Net for Iterative Signal Recovery: Identical Encoding-Decoding CNN Structure} \label{sec:BCD-Net}

To recover a signal $\mb{x} \in \bbC^N$ from a measurement $\mb{y} \in \bbC^{M}$, we consider the following BCD optimization framework with two block variables $\mb{x}$ and $\mb{z}$:
\be{
\label{sys:recover:block}
\argmin_{\mb{x} \in \bbC^N} \min_{\mb{z} \in \bbC^N}~ f(\mb{x}; \mb{y}) +  \lambda \nm{ \mb{x} - \mb{z} }_2^2 + g(\mb{z}),
}
where $f(\mb{x}; \mb{y})$ is a data fitting term and $\mb{z} \in \bbC^N$ is a signal denoised by the regularizer $g(\mb{z})$. In imaging problems, $f(\mb{x}; \mb{y})$ relates to physical imaging models and noise statistics; e.g., \textit{a)} for image denoising, $f(\mb{x}; \mb{y}) = \| \mb{y} - \mb{x} \|_2^2$ where $\mb{y}$ is the noisy image corrupted by additive white Gaussian noise (AWGN); \textit{b)} for MRI, $f(\mb{x}; \mb{y})  = \| \mb{y} - \mb{P}_{\Omega} \mb{F} \mb{x} \|_2^2$, where $\mb{y}$ is the \textit{k}-space measurement and $\mb{P}_{\Omega} \mb{F}$ is an undersampled Fourier operator with $\Omega \subseteq \{ 1,\ldots, N \}$.
Examples of $g(\mb{z})$ include learned convolutional operators, e.g., convolutional dictionary \cite{Chun&Fessler:18TIP, Chun&Fessler:17SAMPTA} and convolutional analysis operator \cite{Chun&Fessler:18arXiv}.
The BCD-Net incorporates the iteration-wise trained image mapping networks into the BCD algorithmic framework in \R{sys:recover:block}. See Algorithm~\ref{alg:BCD-Net}.

The signal recovery performance of BCD-Net largely depends on the performance of $\mathrm{Mapping}^{(i)} (\mb{x}^{(i)})$ in Algorithm~\ref{alg:BCD-Net}. 
Our goal is to reduce the number of layers by designing better image mapping networks that achieve more accurate image recovery. 
Motivated by designing $g(\mb{z})$ with the learned convolutional operators \cite{Chun&Fessler:18TIP, Chun&Fessler:17SAMPTA, Chun&Fessler:18arXiv}, we are particularly interested in the following image mapping CNN using the identical encoding-decoding structures:
\begingroup
\setlength{\thinmuskip}{1.5mu}
\setlength{\medmuskip}{2mu plus 1mu minus 2mu}
\setlength{\thickmuskip}{2.5mu plus 2.5mu}
\be{
\label{sys:cnn:map}
\mathrm{Mapping}^{(i)} (\mb{x}^{(i)}) := \sum_{k=1}^K \mb{S} \mb{d}_k^{(i)} \circledast \cT_{\alpha^{(i)}_k} \!\! \left( \big( \mb{d}_k^{(i)} \big)^{\!*} \circledast \mb{x}^{(i)} \right), \tag{P0}
}
\endgroup
for $i=1,\ldots,N_{\text{Nets}}$, where $\mathrm{Mapping}^{(i)} (\cdot)$ denotes the trained mapping network at the $i\rth$ layer and $N_{\text{Nets}}$ is the number of layers in BCD-Net. Here, $\mb{d}_k^{(i)} \in \bbC^{R}$ denotes the $k\rth$ filter at the $i\rth$ layer, the soft thresholding operator $\cT_{\mb{a}} (\mb{x}) : \bbC^{N} \rightarrow \bbC^{N}$ is defined by
\be{
\label{def:softTh}
( \cT_{\mb{a}} (\mb{x} ) )_n := \left\{ \begin{array}{cc} x_n - a_n \cdot \sgn(x_n), & | x_n | > a_n, \\ 0, & | x_n | \leq a_n, \end{array} \right.
}
for $j = 1,\ldots,N$, $\sgn(\cdot)$ is the (real or complex) sign function, and $\alpha_k^{(i)} \in \bbR$ denotes the $k\rth$ thresholding value at the $i\rth$ layer,
$\mb{S} \in \bbC^{R \times R}$ flips a column vector in the vertical direction (e.g., it rotates 2D filters by $180^\circ$), 
$R$ is the size of filters, $K$ is the number of filters, and $( {\cdot} )^*$ indicates complex conjugate.

In the distinct encoding-decoding structure \cite[(2)]{Ravishankar&Chun&Fessler:17Asilomar}, the decoding filters evolved significantly during training while the encoding filters changed very little.
In the proposed mapping structure \R{sys:cnn:map}, we use same filters both in encoders and decoders to avoid this concern, i.e., we expect that \R{sys:cnn:map} can capture rich information of training data both in encoders and decoders---see Fig.~\ref{fig:filter} later.

\section{Training the Proposed BCD-Net} \label{sec:training}

Reformulating the convolutional operators in \R{sys:cnn:map} with a local approach (i.e., patch-based method) \cite[\protect\S S.\Romnum{1}]{Chun&Fessler:18arXiv}, this section proposes an algorithm for training image mapping CNN \R{sys:cnn:map} in BCD-Net (see Algorithm~\ref{alg:BCD-Net}). 
The training process requires $L$ high-quality training images, $\{ \mb{x}^{\text{train}}_l : l=1,\ldots L \}$, and $L$ training measurements simulated via imaging physics considered by $f(\mb{x}; \mb{y})$ in \R{sys:recover:block}, $\{ \mb{y}^{\text{train}}_l : l=1,\ldots,L \}$. At the $i\rth$ layer, we train $\mathrm{Mapping}^{(i)}$ in \R{sys:cnn:map} as follows:
\ea{
\label{train:cnn:denoiser}
\left\{ \mb{D}^{(i)}, \alphabf^{(i)} \right\} = \argmin_{\{ \mb{D}, \alphabf \}} &~ \nm{ \mb{X}^{(i)}_{\text{train}} - \mb{D} \cT_{\alphabf} \! \left( \mb{D}^H \mb{X}^{(i-1)} \right)  }_F^2, 
\nn \\
\mathrm{subj.~to} &~ \| \mb{d}_k \|_2 \leq 1, \quad k=1,\ldots,K. \tag{P1}
}
where $\mb{X}^{(i)}_{\text{train}}, \mb{X}^{(i-1)} \in \bbC^{R \times N'}$ are training data matrices in which columns correspond to $N'$ patches randomly extracted from $\{ \mb{x}_l^{\text{train}} \}$ and $\{  \mb{x}_l^{(i-1)} \}$, respectively, $\{ \mb{x}_l^{(i-1)} : l=1,\ldots,L \}$ is a set of $L$ images recovered at the $(i\!-\!1)\rth$ layer, the filter matrix $\mb{D} \in \bbC^{R \times K}$ is defined by $\mb{D} := \left[ \mb{d}_1, \cdots, \mb{d}_K \right]$, and $\mb{\alphabf}^{(i)} \in \bbR^K$ is a vector consisting of $K$ thresholding values.
Algorithm~\ref{alg:BCD-Net:training} summarizes the training procedure.

To train $\mathrm{Mapping}^{(i)}$ via \R{train:cnn:denoiser}, we update $2 K$ blocks sequentially; at the $k\rth$ block, we alternatively update the $k\rth$ filter and thresholding value---$\mb{d}_k^{(i)}$ and $\alpha_k^{(i)}$, respectively. \R{train:cnn:denoiser} can be decomposed as  $2K$ $\{ \mb{d}_k, \alpha_k \}$-update problems \cite{Ravishankar&Chun&Fessler:17Asilomar}:
\ea{
\label{sys:decomp}
\left\{ \mb{d}_k^{(i)}, \alpha_k^{(i)} \right\} = \argmin_{\{ \mb{d}_k, \alpha_k \}} &~ \nm{ \mb{E}_k^{(i)} - \mb{d}_k \cT_{\alpha_k} \! \left( \mb{d}_k^H \mb{X}^{(i-1)} \right)  }_F^2, 
\nn \\
\mathrm{subj.~to} &~ \| \mb{d}_k \|_2 \leq 1,
}
where $\mb{E}_k^{(i)} := \mb{X}^{(i)}_{\text{train}} - \sum_{k' \neq k} \mb{d}_{k'} \cT_{\alpha_{k'}} \! \left( \mb{d}_{k'}^H \widetilde{\mb{X}}^{(i)} \right)$. 
To solve \R{sys:decomp}, we alternatively update $\mb{d}_k^{(i)}$ and $\alpha_k^{(i)}$.

\subsection{$k\rth$ Thresholding Value Update}

Using the current estimates of $\mb{d}_k$, the $k\rth$ thresholding value $\alpha_k$ is updated by subgradient descent method with backtracking line search (for step sizes) \cite[\S 2]{Ravishankar&Chun&Fessler:17Asilomar}.

\begin{algorithm}[pt!]
\caption{Training BCD-Net}
\label{alg:BCD-Net:training}

\begin{algorithmic}
\REQUIRE $\{ \mb{x}^{\text{train}}_l, \mb{x}^{(0)}_l, \mb{y}^{\text{train}}_l : l=1,\ldots,L \}$, $\lambda > 0$

\FOR{$i = 0,\ldots,N_{\text{Nets}}\!-\!1$}

\STATE Train $\mathrm{Mapping}^{(i+1)}$ \R{sys:cnn:map} via \R{train:cnn:denoiser} using $\{ \mb{x}^{\text{train}}_l, \mb{x}^{(i)}_{l} \}$

\FOR{$l = 1,\ldots,L$}

\STATE $\displaystyle \mb{z}^{(i+1)}_l = \mathrm{Mapping}^{(i+1)} \big( \mb{x}^{(i)}_l \big)$
\STATE $\displaystyle \mb{x}^{(i+1)}_l = \argmin_{\mb{x}_l \in \bbC^N} f(\mb{x}_l; \mb{y}^{\text{train}}_l) +  \lambda \big\| \mb{x}_l - \mb{z}^{(i+1)}_l \big\|_2^2$

\ENDFOR

\ENDFOR

\end{algorithmic}
\end{algorithm}

\subsection{$k\rth$ Filter Update} \label{sec:filter:ADMM}

Using the current update of $\alpha_k$, we apply ADMM \cite[\S 3.1.1]{Boyd&Parikh&Chu&Peleato&Eckstein:11FTML} to update the $k\rth$ filter $\mb{d}_k$. 
We update the $k\rth$ filter $\mb{d}_k$ by augmenting \R{sys:decomp} with auxiliary variables (dropping the filter indices $k$ and layer indices $(i), (i-1)$ for simplicity):
\eas{
\mb{d}_k = \argmin_{\mb{d}} &~ \nm{ \mb{E} - \mb{d} \cT_{\alpha} ( \mb{v}^H ) }_F^2
\nn \\
\mathrm{subj.~to} &~ \| \mb{d} \|_2 \leq 1, \quad \mb{v} = \mb{X}^H \mb{d}.
}
The cost function above has the corresponding augmented Lagrangian:
\eas{
\cL(\mb{d}, \mb{v}) =& \, \frac{1}{2} \nm{ \mb{E} - \mb{d} \cT_\alpha ( \mb{v}^H ) }_F^2 + \frac{\rho}{2} \nm{ \mb{X}^H \mb{d} - \mb{v} +  \mb{u}  }_2^2 
\\
& + \cI_{\{ \mb{d} : \| \mb{d} \|_2 \leq 1 \}} (\mb{d}),
}
where $\cI_{S} (\mb{x})$ is the indicator function defined by $\cI_S (\mb{x}) = 0$, if $\mb{x} \in S$, and $\cI_S (\mb{x}) = \infty$, otherwise.
We descend/ascend the augmented Lagrangian $\cL(\mb{d}, \mb{v})$, using the following iterative updates of the primal, auxiliary, dual variables---$\mb{d}$, $\mb{v}$, and $\mb{u}$, respectively:
\begingroup
\allowdisplaybreaks
\begin{subequations}
\ea{
\mb{v}^{(j+1)} & = \argmin_{\mb{v}} \, \frac{1}{2} \nm{ \mb{E} - \mb{d}^{(j)} \cT_{\alpha} ( \mb{v} )  }_F^2 
\nn \\
& \hspace{4.6em} + \frac{\rho^{(j)}}{2} \nm{ \mb{v} - \big( \mb{X}^H \mb{d}^{(j)} + \mb{u}^{(j)} \big) }_2^2;
\label{eq:v} 
\\
\mb{d}^{(j+1)} & = \argmin_{\mb{d}} \, \frac{1}{2} \nm{ \mb{E} - \mb{d} \cT_{\alpha} \Big( (\mb{v}^{(j+1)})^H \Big) }_F^2 
\nn \\
& \hspace{4.6em} + \frac{\rho^{(j)}}{2} \nm{ \mb{X}^H \mb{d}  - \big( \mb{v}^{(j+1)} - \mb{u}^{(j)} \big) }_2^2
\nn \\
& \hspace{4.6em} + \cI_{\{ \mb{d} : \| \mb{d} \|_2 = 1 \}} (\mb{d});
\label{eq:d} 
\\
\mb{u}^{(j+1)} &= \mb{u}^{(j)} +  \mb{X}^H \mb{d}^{(j+1)} - \mb{v}^{(j+1)},
\label{eq:u} 
}
\end{subequations}
\endgroup
where the ADMM parameters $\{ \rho^{(j+1)} \}$ are fixed or change based on some adaptive rules, e.g., residual balancing \cite[\S 3.4.1]{Boyd&Parikh&Chu&Peleato&Eckstein:11FTML}.

We first consider problem \R{eq:d}. Rewrite \R{eq:d} as follows:
\ea{
\label{eq:d:re}
\mb{d}^{(j+1)} = \argmin_{\mb{d}} &~ \frac{\rho^{(j)}}{2} \mb{d}^H \mb{X} \mb{X}^H \mb{d} - \mathrm{Re} \! \left\{ \mb{d}^H \left( \mb{E}  \cT_{\alpha} ( \mb{v}^{(j+1)} ) \right. \right.
\nn \\
& \left. \left. + \rho^{(j)} \mb{X} \big( \mb{v}^{(j+1)} - \mb{u}^{(j)} \big) \right) \right\}
\nn \\
\mathrm{subj.~to} &~ \| \mb{d} \|_2 \leq 1,
}
by $\| \mb{d} \cT_{\alpha} ( \mb{v}^H ) \|_F^2 = c \cdot \| \cT_{\alpha} ( \mb{v} ) \|_2^2$ and $\tr (  \cT_{\alpha} ( \mb{v}^H )^H \mb{d}^H \mb{E} ) = \mb{d}^H \mb{E} \cT_{\alpha} ( \mb{v} )$, using $\cT_{\alpha} ( \mb{v}^H )^H  \!=\! \cT_{\alpha} ( \mb{v} )$ and $\| \mb{d} \|_2^2 \!=\! c \!\leq\! 1$.
We apply an accelerated Newton's method to efficiently obtain the optimal solution to problem \R{eq:d:re} \cite[\S \Romnum{4}-A3]{Chun&Fessler:18arXiv}, \cite[\S \Romnum{4}--\Romnum{5}-A2]{Chun&Fessler:18TIP}, considering that  \R{eq:d:re} is a (convex) quadratically constrained quadratic program.
The closed form solution in \cite[(4)]{Ravishankar&Chun&Fessler:17Asilomar} is not applicable for solving \R{eq:d}, because of an additional quadratic term, e.g., the second term in \R{eq:d}.

For problem \R{eq:v}, we first rewrite \R{eq:v} as 
\ea{
\label{eq:v:re}
\mb{v}^{(j+1)} = \argmin_{\mb{v}} &~ \frac{1}{2} \nm{ \cT_\alpha ( \mb{v} ) - c^{-1} \cdot \mb{E}^H \mb{d}^{(j)} }_2^2 
\nn \\
&\, + \frac{\rho^{(j)}}{2 c} \nm{ \mb{v} - ( \mb{X}^H \mb{d}^{(j)} + \mb{u}^{(j)} ) }_2^2
}
by applying the reformulation tricks used in \R{eq:d:re}.
Using the separability of \R{eq:v:re}, we solve the following element-wise optimization problems:
\be{
\label{eq:v:re:elem}
v^{(j+1)}_n = \argmin_{v_n} \frac{1}{2} \left| \cT_\alpha (v_n) - g_n^{(j)}  \right|^2 + \frac{\rho^{(j)}}{2c} \left| v_n - h_n^{(j)} \right|^2,
}
for $n \!=\! 1,\ldots,N'$, where $\mb{g}^{(j)} \!:=\! c^{-1} \mb{E}^H \mb{d}^{(j)}$ and $\mb{h}^{(j)} \!:=\!  \mb{X}^H \mb{d}^{(j)} + \mb{u}^{(j)}$. 
We solve \R{eq:v:re:elem} by subgradient descent method with backtracking line search: \textit{a)} for the real-valued problem, we apply Lemma~\ref{l:real}, and \textit{b)} for the complex-valued problem, we apply Lemma~\ref{l:complex}.

\lem{\label{l:real}
The gradient of $f(v) = \frac{1}{2} ( \cT_{\alpha} (v) - g )^2 + \frac{\rho}{2} ( v - h )^2$ is given by
\bes{
\partial f(v)/\partial v  = ( \cT_{\alpha}(v) - g) \cdot 1_{|v| > \alpha} + \rho (v - h),
}
where $v, g, h \in \bbR$ and $\alpha, \rho > 0$.
}

\lem{\label{l:complex}
The gradient of $f(v) = \frac{1}{2} | \cT_{\alpha} (v) - g |^2 + \frac{\rho}{2} | v - h |^2$ is given by
\eas{
\partial f(v)/\partial v =\, & ( \cT_{\alpha}(v) - g ) \cdot 1_{|v| > \alpha} + \rho (v - h)
\nn \\
& +  \I \alpha |v|^{-3} \cdot  \Im \{ v \cdot \left( \cT_{\alpha} (v) - g \right)^* \}  \cdot 1_{|v| > \alpha},
}
where $v, g, h \in \bbC$ and $\alpha, \rho > 0$.
Distinct from the index $i$, we denote the imaginary unit by $\I$.
}
\prf{\renewcommand{\qedsymbol}{}
See Appendix.  
}

Note that Lemma~\ref{l:real} is a special case of Lemma~\ref{l:complex}: when imaginary components of $\{ v,g,h \}$ in Lemma~\ref{l:complex} vanish, the gradient $\frac{\partial f(v)}{\partial v}$ in Lemma~\ref{l:complex} becomes that in Lemma~\ref{l:real}.

\section{Results and Discussion}

\begin{figure}[!t]
\small\addtolength{\tabcolsep}{-5pt}
\centering

\begin{tabular}{cccc}

\includegraphics[scale=0.28, trim=4.1em 4.1em 2.8em 2.6em, clip]{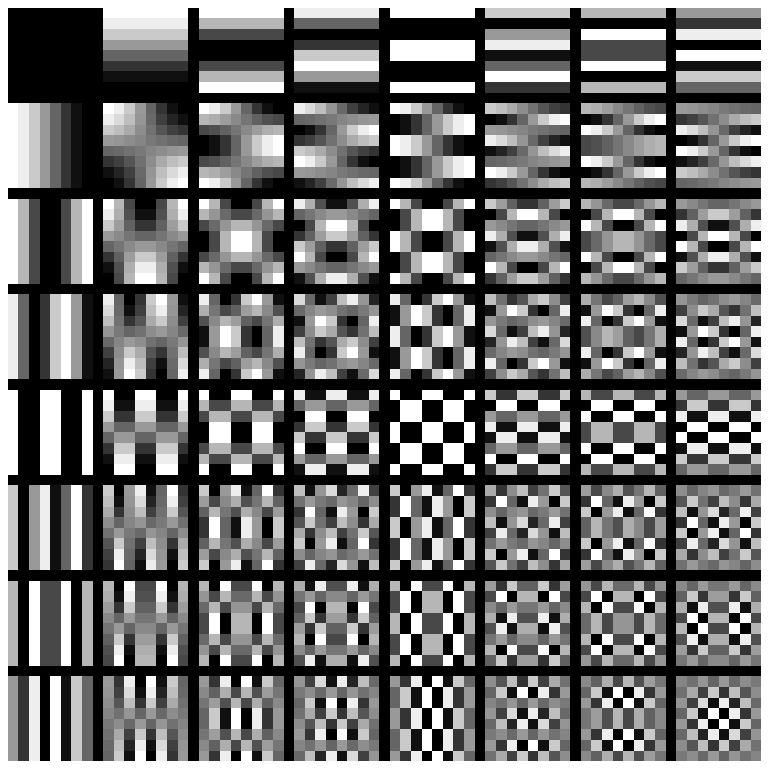} &
\includegraphics[scale=0.28, trim=4.1em 4.1em 2.8em 2.6em, clip]{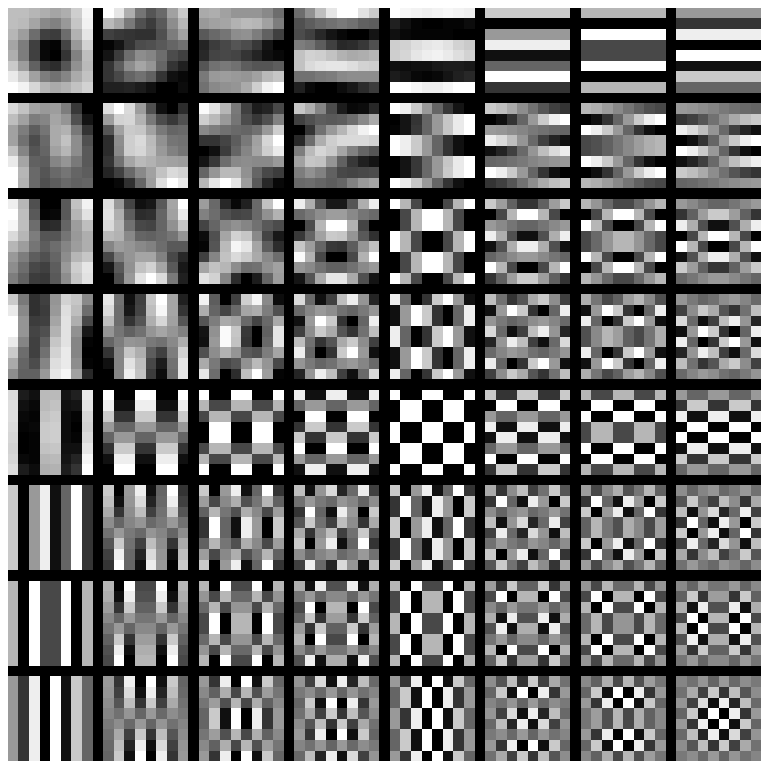} &
\includegraphics[scale=0.28, trim=4.1em 4.1em 2.8em 2.6em, clip]{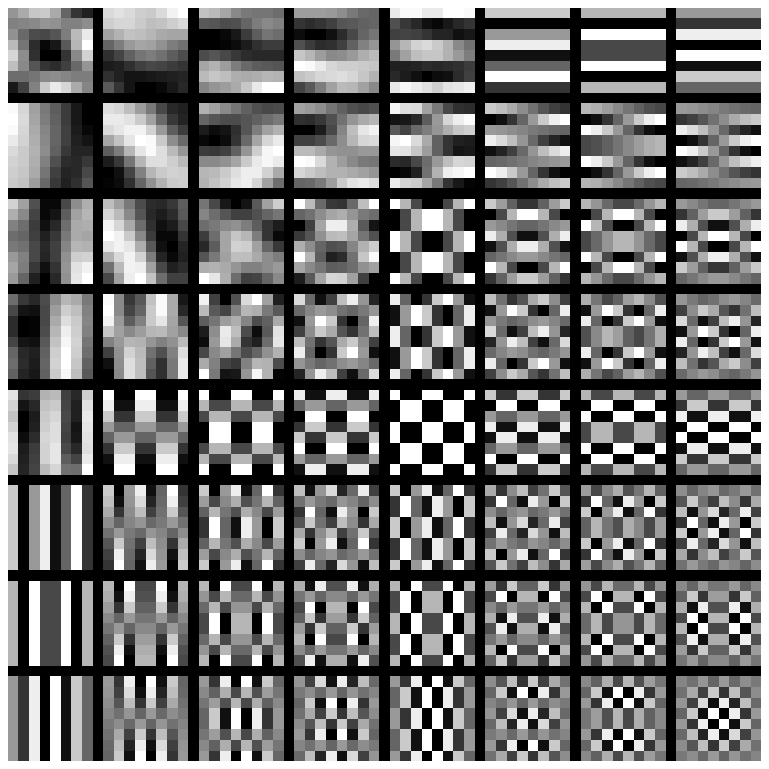}  &
\includegraphics[scale=0.28, trim=4.1em 4.1em 2.8em 2.6em, clip]{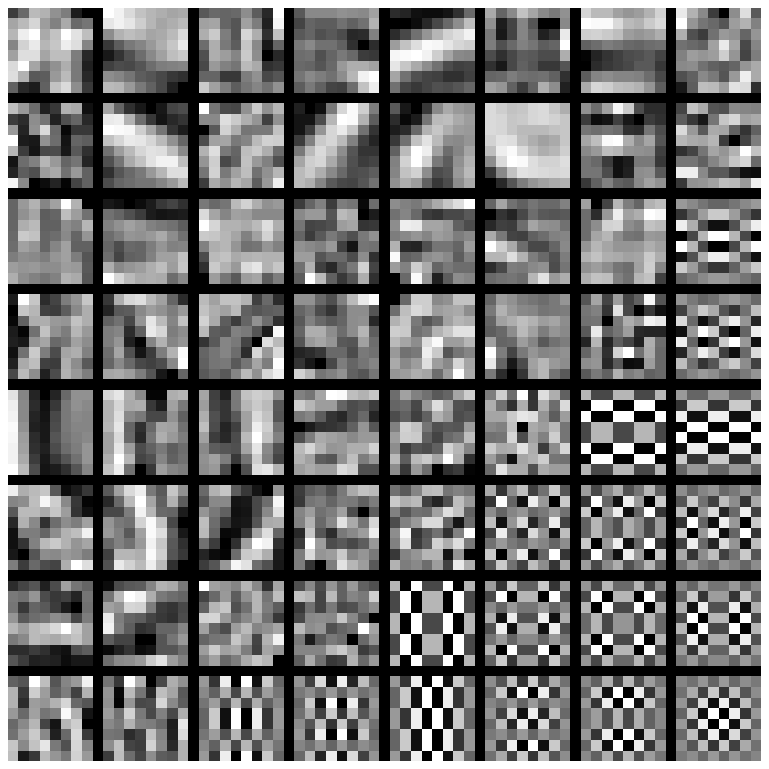} \vspace{-0.25em} \\

{\small (a)} & {\small (b)} & {\small (c)} & {\small (d)}

\end{tabular}

\vspace{-0.25em}
\caption{Examples of trained filters for BCD-Net (MR image reconstruction; real components of complex-valued filters trained at the $10\rth$ layer). 
(a) Initial filters (discrete cosine transform).
(b-c) Encoding and decoding filters trained by the image mapping module in \cite{Ravishankar&Chun&Fessler:17Asilomar}.
(d) Filters trained with the proposed image mapping module using the identical encoding-decoding CNN structure.
}
\vspace{-1em}
\label{fig:filter}
\end{figure}

\begin{figure*}[!t]
\centering
\small\addtolength{\tabcolsep}{-7.5pt}
\renewcommand{\arraystretch}{1}

\begin{tabular}{cccc}
&
\small{\specialcell{(1) Image denoising \\ ($\sigma = 20$)}}
&
\small{\specialcell{(2) Image denoising \\ ($\sigma = 30$)}}
&
\small{\specialcell{(3) MR image reconstruction \\ (10\% sampling for $256\!\times\!256$ res.)}}
\\

\raisebox{-.5\height}{\begin{turn}{+90} \small{(a) Training @ $1\mathrm{st}$ layer} \end{turn}}~ 
&
\raisebox{-.5\height}{
	\includegraphics[scale=0.54, trim=0.6em 0em 2em 0.2em, clip]{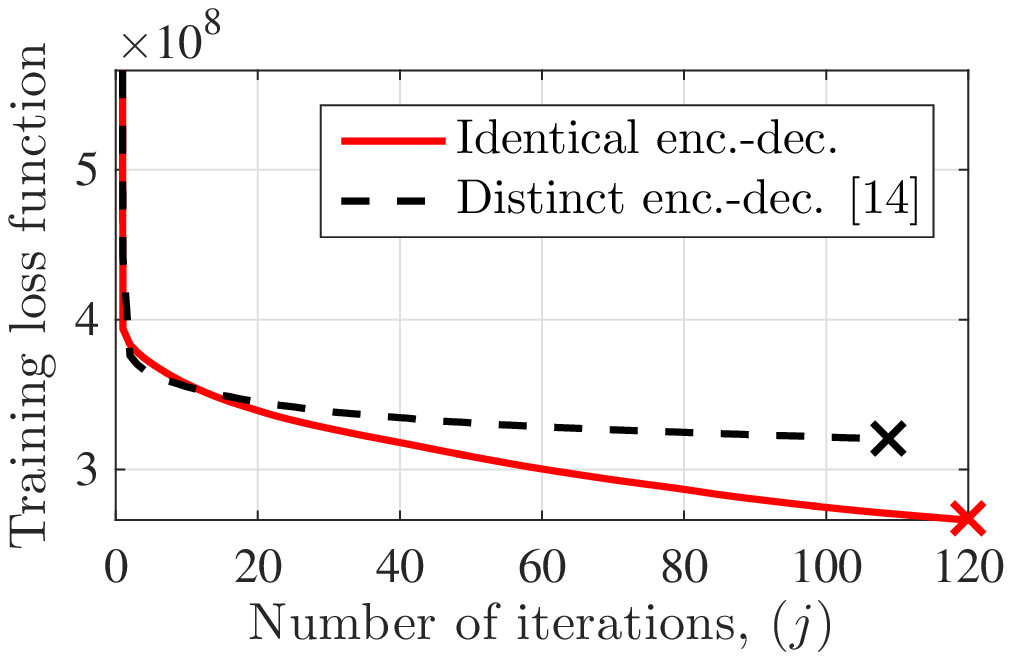}
	}
&
\raisebox{-.5\height}{
	\includegraphics[scale=0.54, trim=0.6em 0em 2em 0.2em, clip]{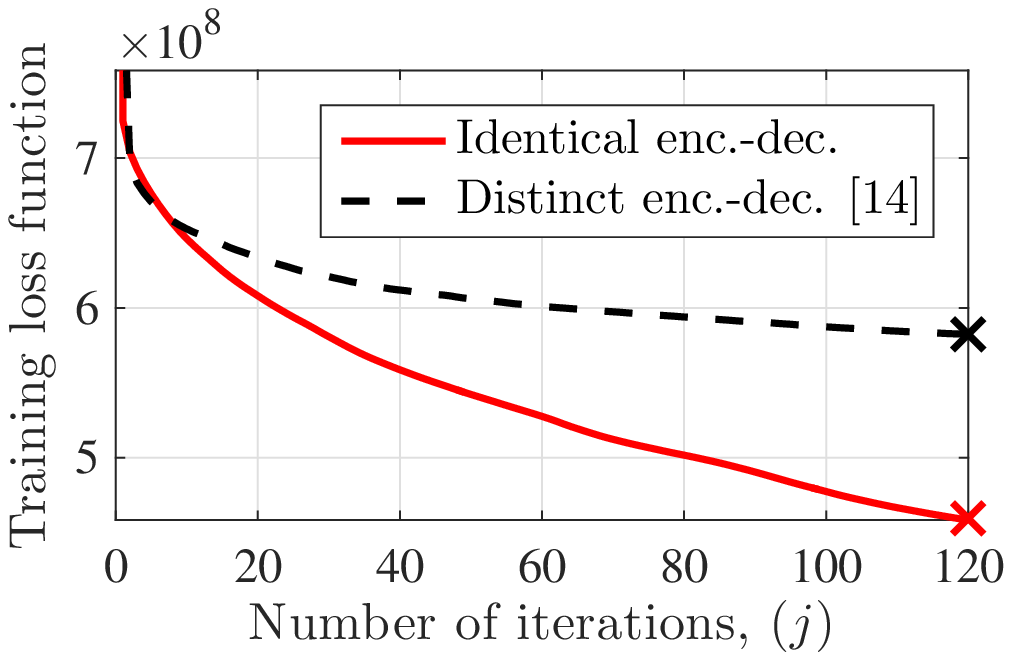}
	}
&
\raisebox{-.5\height}{
	\includegraphics[scale=0.54, trim=0em 0em 2em 0.2em, clip]{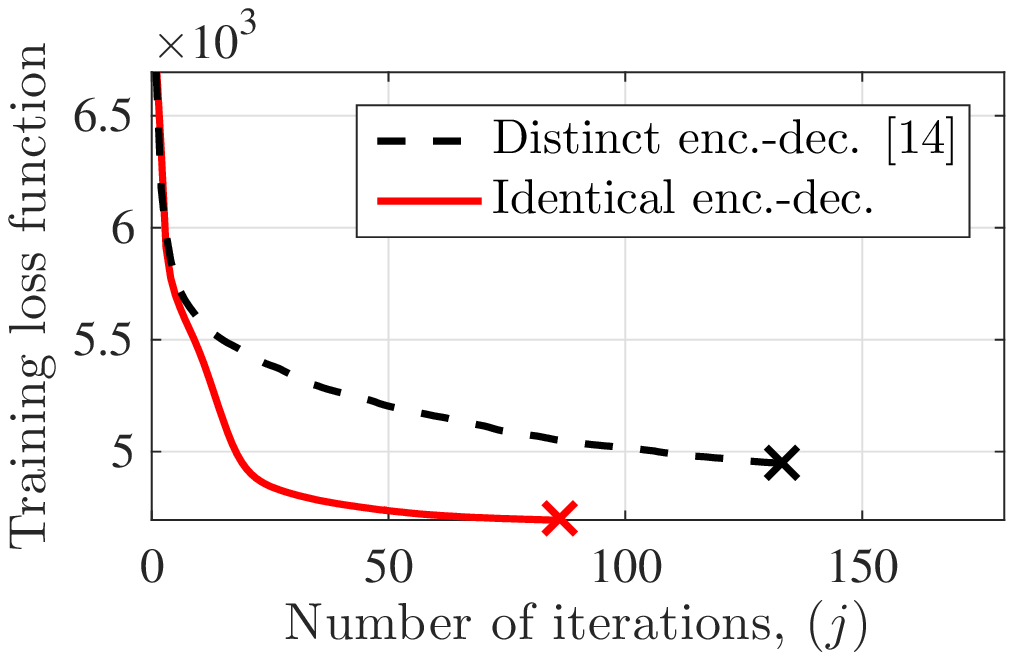}
	}
\\

\raisebox{-.5\height}{\begin{turn}{+90} \small{(b) Testing} \end{turn}}~ 
&
\raisebox{-.5\height}{
	\includegraphics[scale=0.53, trim=0.2em 0em 2em 0.5em, clip]{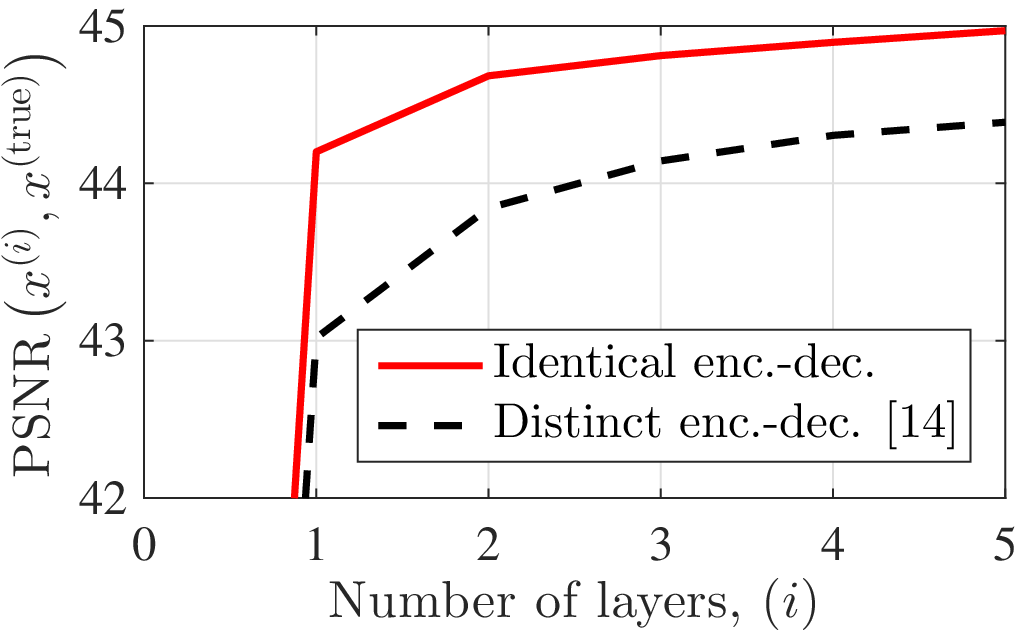}
	}
&
\raisebox{-.5\height}{
	\includegraphics[scale=0.53, trim=0.2em 0em 2em 0.5em, clip]{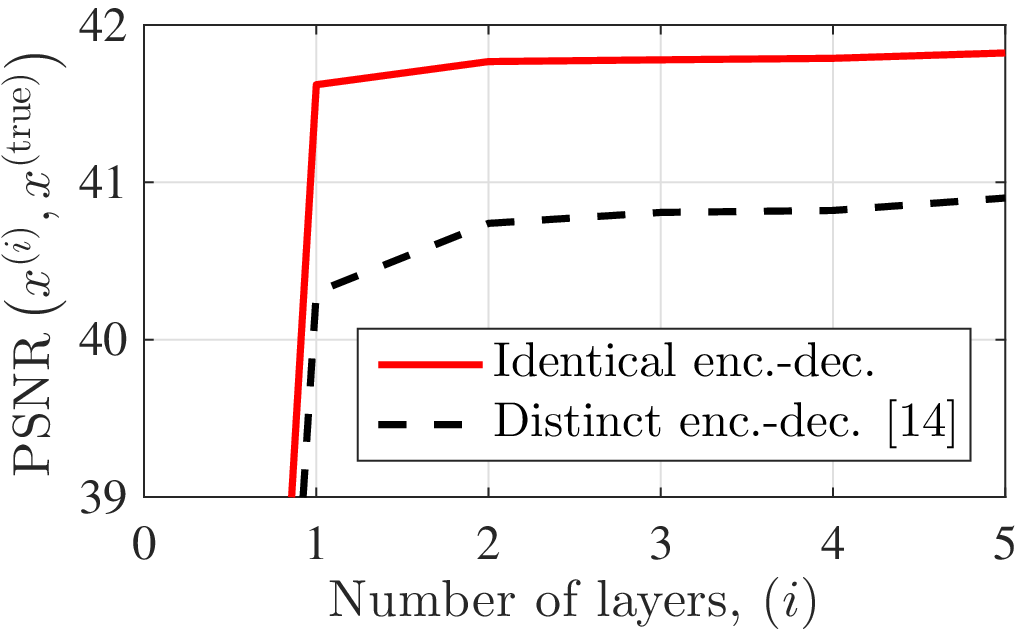}
	}
&
\raisebox{-.5\height}{
	\includegraphics[scale=0.53, trim=-0.6em 0em 2em 0.5em, clip]{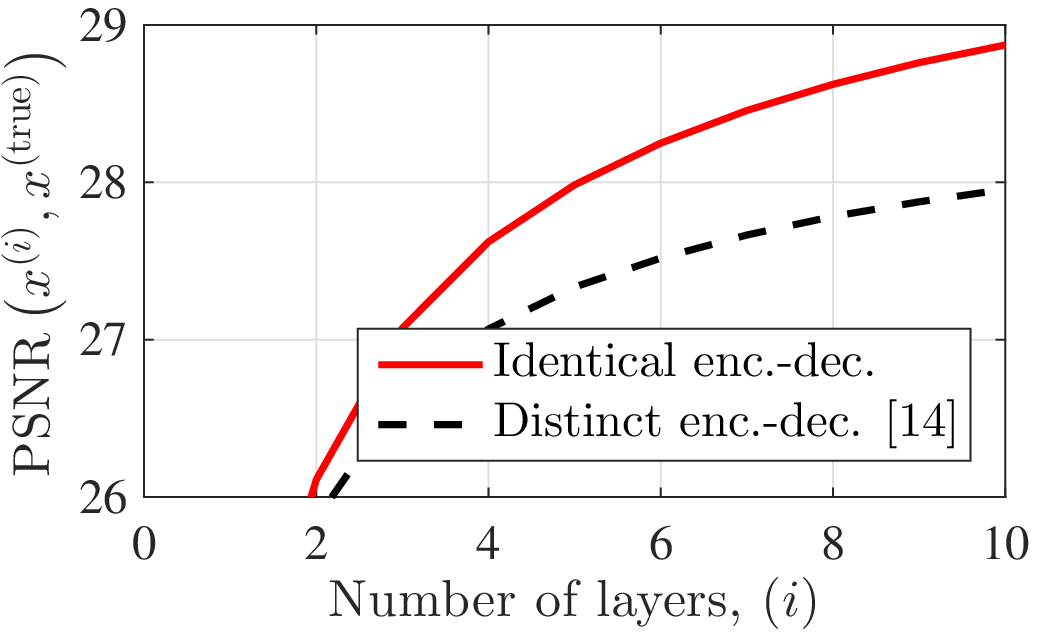}
	}
\end{tabular}

\vspace{-0.25em}
\caption{Comparison of image denoising accuracy for different encoding-decoding structures in BCD-Net and noise levels
(the cross mark \text{\sffamily x} in (a) denotes a termination point). PSNR gaps between the two BCD-Nets are give as follows: (1) $[0.58, 1.18]$dB; (2) $[0.92,1,32]$dB; and (3) $[0.16,0.92]$dB}
\label{fig:train&test}
\end{figure*}

\begin{figure*}[!t]
\centering
\small\addtolength{\tabcolsep}{-7.5pt}
\renewcommand{\arraystretch}{0.1}

    \begin{tabular}{ccccc}
    
            {} & {(a) Full sampling} & \small{(b) Wavelet \& TV \cite{Lustig&Dohono&Pauly:09MRM}} & \small{(c) BCD-Net \cite{Ravishankar&Chun&Fessler:17Asilomar}} & \small{(d) Proposed BCD-Net} \\

       \raisebox{-.5\height}{\begin{turn}{+90} \small{$10$\% sampling for $256\!\times\!256$ res.} \end{turn}}~ &
        \raisebox{-.5\height}{
        \begin{tikzpicture}
            \begin{scope}[spy using outlines={rectangle,yellow,magnification=1.7,size=14mm, connect spies}]
            	\node {\includegraphics[scale=0.45]{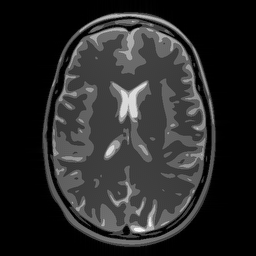}};
		\spy on (0.05,-0.1) in node(a) [left] at (2.1,-2.1);
                 \spy on (-0.8,-0.6) in node(b) [left] at (-0.7,-2.1);
            \end{scope}
            \begin{scope}[only in spy node magn 1.75]
                \draw [red, line width=0.1mm] (-0.05,0.2) circle [radius=0.1];
 	        \draw [red, line width=0.1mm] (0,-0.2) circle [radius=0.1];
	        \draw [red, line width=0.1mm] (1.55,-0.04) circle [radius=0.16];
            \end{scope}
        \end{tikzpicture}} &
        \raisebox{-.5\height}{
        \begin{tikzpicture}
            \begin{scope}[spy using outlines={rectangle,yellow,magnification=1.7,size=14mm, connect spies}]
            	\node {\includegraphics[scale=0.45]{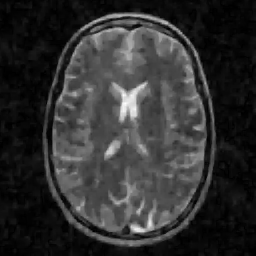}};
		\spy on (0.05,-0.1) in node [left] at (2.1,-2.1);
                 \spy on (-0.8,-0.6) in node [left] at (-0.7,-2.1);
		\node [white] at (0.85,1.85) {\scriptsize $\mathrm{PSNR} = \mb{23.81}$ dB};
            \end{scope}
            \begin{scope}[only in spy node magn 1.75]
                \draw [red, line width=0.1mm] (-0.05,0.2) circle [radius=0.1];
 	        \draw [red, line width=0.1mm] (0,-0.2) circle [radius=0.1];
	        \draw [red, line width=0.1mm] (1.55,-0.04) circle [radius=0.16];
            \end{scope}
        \end{tikzpicture}} &
        \raisebox{-.5\height}{
        \begin{tikzpicture}
            \begin{scope}[spy using outlines={rectangle,yellow,magnification=1.7,size=14mm, connect spies}]
            	\node {\includegraphics[scale=0.45]{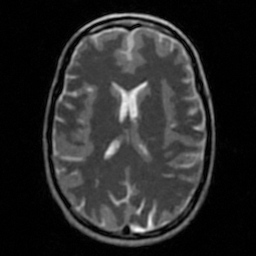}};
		\spy on (0.05,-0.1) in node [left] at (2.1,-2.1);
                 \spy on (-0.8,-0.6) in node [left] at (-0.7,-2.1);
                \node [white] at (0.85,1.85) {\scriptsize $\mathrm{PSNR} = \mb{27.96}$ dB};
            \end{scope}
            \begin{scope}[only in spy node magn 1.75]
                \draw [red, line width=0.1mm] (-0.05,0.2) circle [radius=0.1];
 	        \draw [red, line width=0.1mm] (0,-0.2) circle [radius=0.1];
	        \draw [red, line width=0.1mm] (1.55,-0.04) circle [radius=0.16];
            \end{scope}
        \end{tikzpicture}} &
        \raisebox{-.5\height}{
        \begin{tikzpicture}
            \begin{scope}[spy using outlines={rectangle,yellow,magnification=1.7,size=14mm, connect spies}]
            	\node {\includegraphics[scale=0.45]{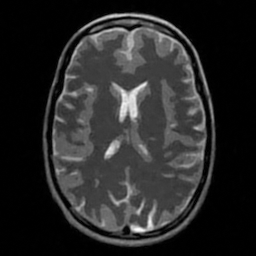}};
                 \spy on (0.05,-0.1) in node [left] at (2.1,-2.1);
                 \spy on (-0.8,-0.6) in node [left] at (-0.7,-2.1);
                \node [white] at (0.85,1.85) {\scriptsize \color{yellow}{$\mathrm{PSNR} = \mb{28.87}$ dB}};
            \end{scope}
            \begin{scope}[only in spy node magn 1.75]
                \draw [red, line width=0.1mm] (-0.05,0.2) circle [radius=0.1];
 	        \draw [red, line width=0.1mm] (0,-0.2) circle [radius=0.1];
	        \draw [red, line width=0.1mm] (1.55,-0.04) circle [radius=0.16];
            \end{scope}
        \end{tikzpicture}} 

    \end{tabular}

\vspace{-0.25em}
\caption{Comparison of reconstructed MR images from BCD-Nets using different image mapping CNN modules (at the $10\rth$ layer) and Sparse MRI reconstruction \cite[built-in parameter setting]{Lustig&Dohono&Pauly:09MRM}. (d) Compared to (b) (non-trained) Sparse MRI reconstruction, the proposed BCD-Net improves PSNR by $5$dB.}
\label{fig:MRIrecon}
\vspace{-1em}
\end{figure*}
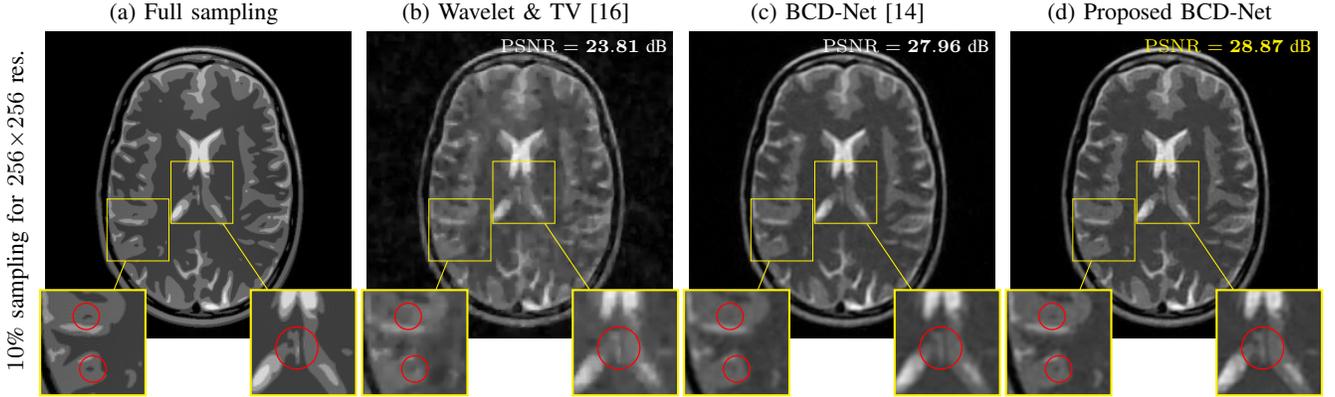

\subsection{Experimental Setup}

\subsubsection{Imaging and Image Recovery}

For image denoising experiments, we contaminated five slices of XCAT phantom \cite{Segars&etal:08MP} by (zero-mean) AWGN with large standard deviation $\sigma  \!\approx\! 135, 202$ HU (that corresponds to $\sigma = 20,30$ for natural images within $[0,255]$); we used four of them for training and the remaining one for testing. 
For MR image reconstruction experiments, we simulated two extremely undersampled ($10$\%) \textit{k}-space data sets with the optimal multi-level sampling in compressed sensing \cite{Adcock&etal:16FMS, Chun&Adcock:17TIT} and field-of-view of $28 \!\times\! 28$cm on the $256 \!\times\! 256$ cartesian grid, while avoiding an inverse crime with two $768\!\times\!768$ complex-valued phantoms \cite{Guerquin-Kern&Lejeune&Pruessmann&Unser:12TMI}; we used one for training and another for testing.
We set the regularization parameter $\lambda$ as follows: for image denoising, $\lambda \!=\! 10/\sigma'$, where $\sigma'$ is scaled $\sigma$ by considering the maximum value of XCAT phantom; for MR image reconstruction, $\lambda \!=\! 10^6$ (the same values were used for training BCD-Nets).
We evaluated the quality of recovered images by peak SNR (PSNR).

\subsubsection{Training BCD-Nets}

We trained $K \!=\! 64$ filters of size $R \!=\! 8 \!\times\! 8$, with $20,000$ randomly extracted image patches in each layer.
For training BCD-Net using the distinct encoding-decoding CNN \cite[(2)]{Ravishankar&Chun&Fessler:17Asilomar}, we used the parameter set (including the number of subgradient descent iterations, filter initialization, initial thresholding values, etc.) in \cite{Ravishankar&Chun&Fessler:17Asilomar}.
For training the proposed BCD-Net using the identical encoding-decoding CNN \R{sys:cnn:map}, we used the parameter set in \cite{Ravishankar&Chun&Fessler:17Asilomar} as the default.
The parameters related to ADMM in Section~\ref{sec:filter:ADMM} are given as follows: we used $4$ ADMM iterations and $4$ inner subgradient descent iterations for updating $\mb{v}^{(j+1)}$ in \R{eq:v}; and we applied the residual balancing scheme \cite[\S 3.4.1]{Boyd&Parikh&Chu&Peleato&Eckstein:11FTML} to adaptively control $\{ \rho^{(j)} \!:\! \forall j \!\neq\! 0 \}$ ($\rho^{(0)} \!=\! 1$).
We terminated the iterations of training each $\mathrm{Mapping}^{(i+1)} (\mb{x}^{(i)})$, if \textit{a)} the relative difference stopping criterion (e.g., \cite[(44)]{Chun&Fessler:18TIP}) is met or \textit{b)} the training costs (e.g., \R{train:cnn:denoiser}) increase, before reaching the maximum number of iterations.
We set the relative difference tolerance as $2\!\times\!10^{-3}$; and the maximum number of iterations to $120$ and $180$ for image denoising and MR image reconstruction, respectively.

\subsection{Relation Between Filter Richness and Image Mapping Performance}

The filters trained with the identical encoding-decoding CNN structure capture diverse features of training data. The rich features captured in filters are useful for better image mapping between corrupted and noiseless images, i.e., lower cost value in \R{train:cnn:denoiser}. See Figs.~\ref{fig:filter}--\ref{fig:train&test}(a).
On the other hand, the filters trained with the distinct encoding-coding structure \cite[(2)]{Ravishankar&Chun&Fessler:17Asilomar} do not sufficiently capture features of training data. 
See Fig.~\ref{fig:filter}(b)--(c). (The result in Fig.~\ref{fig:filter}(b) corresponds to that in \cite[Fig.~2, transform rows]{Ravishankar&Chun&Fessler:17Asilomar}.)

\subsection{Application of Trained BCD-Nets to Iterative Image Denoising and MRI Image Reconstruction}

Promoting better image mapping between the corrupted and noiseless images in training, the proposed BCD-Net (significantly) improves image recovery accuracy compare to the BCD-Net in \cite{Ravishankar&Chun&Fessler:17Asilomar}.
PSNR gaps between the two BCD-Nets across the layers are given as follows: for denoising low SNR images ($\sigma \!=\! 30$), $[0.92, 1.32]$dB; for extremely undersampled MRI ($10$\% sampling), $[0.16, 0.92]$dB. 
For the extremely undersampled MRI experiment, the PSNR gap increases as we increase the number of layers.
See Fig.~\ref{fig:train&test}(b-3).
Compared to the conventional MR reconstruction using wavelets and TV \cite{Lustig&Dohono&Pauly:09MRM}, the proposed BCD-Net significantly improves reconstruction accuracy only with $10$ layers (or iterations): it improves PNSR by $5$dB---see Fig.~\ref{fig:MRIrecon}.

The drawback using BCD-Nets is that BCD-Nets are only applicable when the imaging models in training and testing are identical; meanwhile, the iterative signal recovery using learned CNNs is free from this limitation \cite{Chun&Fessler:18TIP, Chun&Fessler:18arXiv, Chun&Fessler:17SAMPTA}.

\section{Conclusion}

The proposed BCD-Net achieves accurate image recovery within a reasonable computing time in \dquotes{extreme} computational imaging.
The identical encoding-decoding CNN structure provides better image mapping than the distinct structure in \cite[(2)]{Ravishankar&Chun&Fessler:17Asilomar}, by better capturing rich information of training data.
Future works include deriving closed-form solutions to \R{eq:v} for faster training of the proposed BCD-Net and testing its image mapping performances for more (locally) structured artifacts (e.g., aliasing artifacts caused by radial line or spiral undersampling in MRI, and streak artifacts caused by sparse-view computed tomography).

\section*{Appendix: Proofs for Lemma~\ref{l:complex}} \label{sec:prf:l:complex}

For $| v | > \alpha$, we rewrite $f(v)$ in Lemma~\ref{l:complex} as follows:
\begingroup
\setlength{\thinmuskip}{1.5mu}
\setlength{\medmuskip}{2mu plus 1mu minus 2mu}
\setlength{\thickmuskip}{2.5mu plus 2.5mu}
\eas{
f(v_R, v_I) =\, & \frac{1}{2} \left( v_R - \frac{\alpha v_R}{\sqrt{v_R^2 + v_I^2}} - g_R \right)^2 + \frac{\rho}{2} \left( v_R - h_R \right)^2
\\
& + \frac{1}{2} \left( v_I - \frac{\alpha v_I}{\sqrt{v_R^2 + v_I^2}} - g_I \right)^2 + \frac{\rho}{2} \left( v_I - h_I \right)^2,
}
\endgroup
where $v = v_R + \I v_I$, $g = g_R + \I g_I$, and $h = h_R + \I h_I$.
Taking partial derivatives of $f(v_R, v_I)$ with respect to $v_R$ and $v_I$, we obtain
\begingroup
\setlength{\thinmuskip}{1.5mu}
\setlength{\medmuskip}{2mu plus 1mu minus 2mu}
\setlength{\thickmuskip}{2.5mu plus 2.5mu}
\allowdisplaybreaks
\eas{
\frac{\partial f(v_R, v_I)}{\partial v_R} =&  \left( v_R - \frac{\alpha v_R}{\sqrt{v_R^2 + v_I^2}} - g_R \right) \left( 1 - \frac{\alpha v_I^2}{\left( v_R^2 + v_I^2 \right)^{3/2}} \right)
\\
& + \left( v_I - \frac{\alpha v_I}{\sqrt{v_R^2 + v_I^2}} - g_I \right) \frac{\alpha v_R v_I}{\left( v_R^2 + v_I^2 \right)^{3/2}}
\\
& + \rho \left( v_R - h_R \right),
\nn \\
\frac{\partial f(v_R, v_I)}{\partial v_I} =&  \left( v_I - \frac{\alpha v_I}{\sqrt{v_R^2 + v_I^2}} - g_I \right) \left( 1 - \frac{\alpha v_R^2}{\left( v_R^2 + v_I^2 \right)^{3/2}} \right)
\\
& + \left( v_R - \frac{\alpha v_R}{\sqrt{v_R^2 + v_I^2}} - g_R \right) \frac{\alpha v_R v_I}{\left( v_R^2 + v_I^2 \right)^{3/2}}
\\
& + \rho \left( v_I - h_I \right).
}
\endgroup
Rearraging the results above, we attain
\begingroup
\setlength{\thinmuskip}{1.5mu}
\setlength{\medmuskip}{2mu plus 1mu minus 2mu}
\setlength{\thickmuskip}{2.5mu plus 2.5mu}
\allowdisplaybreaks
\ea{
\label{grad:complex:|v|>alpha}
\frac{\partial f(v)}{\partial v} 
=&\, \frac{\partial f(v_R, v_I)}{\partial v_R} + \I \frac{\partial f(v_R, v_I)}{\partial v_I} 
\nn \\
=&\, \zeta_R (v) \! \left( 1 - \frac{\alpha v_I^2}{\sqrt{v_R^2 + v_I^2}} \right) 
+ \I \zeta_I (v) \! \left( 1 - \frac{\alpha v_R^2}{\sqrt{v_R^2 + v_I^2}} \right)
\nn \\
&+ \frac{\alpha v_R v_I}{ \left( v_R^2 + v_I^2 \right)^{3/2} } \left( \zeta_I (v) + \I \zeta_R (v) \right) 
+ \rho (v - h)
\nn \\
=&\, \zeta (v) + \rho (v - h) 
+ \frac{\alpha v_R v_I}{ \left( v_R^2 + v_I^2 \right)^{3/2} } \left( \zeta_I (v) + \I \zeta_R (v) \right)
\nn \\
& - \frac{ \alpha \zeta_R(v) v_I^2 }{\left( v_R^2 + v_I^2 \right)^{3/2} } - \I \frac{  \alpha \zeta_I(v) v_R^2 }{\left( v_R^2 + v_I^2 \right)^{3/2} }
\nn \\
=&\, \zeta (v) + \rho (v - h) + \frac{\alpha}{|v|^3} \Big( v_I (v_R \zeta_I (v) - v_I \zeta_R (v)) 
\nn \\
& + \I v_R ( v_I \zeta_R(v) -  \zeta_I (v) v_R ) \Big)
\nn \\
=&\, \zeta (v) + \rho (v - h) + \frac{\alpha}{ |v|^3 } (v_I - \I v_R)  (v_R \zeta_I(v) - v_I \zeta_R(v))
\nn \\
=&\, \zeta (v) + \rho (v - h) + \frac{\alpha}{ |v|^3 } \cdot (-\I v) \cdot \left( - \Im \left\{ v \cdot \zeta^* (v) \right\} \right),
}
\endgroup
where $\zeta(v) = v - \sgn(v) - g = \zeta_R (v) + \I \zeta_I (v)$.
For $| v | \leq \alpha$, it is trivial to show that
\begingroup
\setlength{\thinmuskip}{1.5mu}
\setlength{\medmuskip}{2mu plus 1mu minus 2mu}
\setlength{\thickmuskip}{2.5mu plus 2.5mu}
\be{
\label{grad:complex:|v|<=alpha}
\frac{\partial f(v)}{\partial v} = \rho (v - h).
}
\endgroup
Combining \R{grad:complex:|v|>alpha} and \R{grad:complex:|v|<=alpha} completes the proof.

\bibliographystyle{IEEEtran}
\bibliography{IEEEabrv,referenceBibs_Bobby}

\end{document}